\documentclass{amia}
\usepackage{graphicx}
\usepackage[labelfont=bf]{caption}
\usepackage[superscript,nomove]{cite}
\usepackage{color}
\usepackage{subcaption}
\usepackage{hyperref}
\usepackage{cleveref}
\usepackage{multirow}
\usepackage{algorithm}
\usepackage{algpseudocode} 

\usepackage{moreverb} 

\begin{document}

\title{MT-Clinical BERT: Scaling Clinical Information Extraction with Multitask Learning}

\author{Andriy Mulyar and Bridget T. McInnes, Ph.D}

\institutes{
    Virginia Commonwealth University, Richmond, Virginia, United States\\
}

\maketitle

\noindent{\bf Abstract}

\textit{Clinical notes contain an abundance of important but not-readily accessible information about patients. Systems to automatically extract this information rely on large amounts of training data for which their exists limited resources to create. Furthermore, they are developed dis-jointly; meaning that no information can be shared amongst task-specific systems. This bottle-neck unnecessarily complicates practical application, reduces the performance capabilities of each individual solution and associates the engineering debt of managing multiple information extraction systems. We address these challenges by developing Multitask-Clinical BERT \footnote{ \url{https://github.com/AndriyMulyar/multitasking\_transformers}}: a single deep learning model that simultaneously performs eight clinical tasks spanning entity extraction, PHI identification, language entailment and similarity by sharing representations amongst tasks. We find our single system performs competitively with all state-the-art task-specific systems while also benefiting from massive computational benefits at inference.}

\section*{Introduction}
Electronic Health Records (EHR) contain a wealth of actionable patient information in the form of structured fields and unstructured narratives within a patient's clinical note. While structured data such as billing codes provide coarse grained signal pertaining to common conditions or treatments a patient may have experienced, a large quantity of vital information is not directly accessible due to being stored in unstructured, free-text notes. The task of automatically extracting structured information from this free-form text is known as \textit{information extraction} and has been an intensely studied line of research over the past two decades. While the primary objective of information extraction is to gather fine-grained information about patients such as problems experienced, treatments underwent, tests conducted and drugs received, auxiliary tasks such as the automatic identification and subsequent removal of Personal Health Information (PHI) are also of pragmatic interest to the functioning of the health system controlling the EHR.

To support this diverse set of information extraction challenges, several community lead shared tasks have annotated datasets for the construction and evaluation of automated information extraction systems. These include the identification of problems, treatments and tests \cite{uzuner20112010, sun2013evaluating}; the identification of drugs, adverse drug events and drug related information \cite{henry20202018}; and the de-identification of PHI \cite{stubbs2015automated}. While these shared tasks have produced well performing solutions, the resulting systems are disjoint meaning that no information is shared between systems addressing each individual information extraction task. Notably, this means that each task requires a separate engineering effort to solve, narrow technical expertise to construct and disjoint computational resources to apply in clinical practice. Recently, this gap has been narrowed by advances in large scale self-supervised text pre-training \cite{devlin-etal-2019-bert, yang2019xlnet}. This paradigm has resulted in well know language representation systems such as BERT which can easily be adapted to any single domain specific task and achieve state-of-the-art performance. In the clinical space, researchers have similarly leveraged large clinical note repositories such as MIMIC-III \cite{johnson2016mimic} to pre-train Clinical BERT \cite{alsentzer-etal-2019-publicly} instances achieving large performance gains on several clinical NLP related tasks. While well-performing, a single fine-tuned Clinical BERT instance requires significant resources to deploy into a clinical informatics workflow thus limiting it's practical applicability. This fact is amplified by the observation that an isolated 110 million parameter model is required \textit{for each} clinical task; scaling linearly the required hardware resources.

This work introduces Multitask-Clinical-BERT: a \textbf{single}, \textbf{unified} deep learning-based clinical information extraction system that concurrently addresses eight information extraction tasks via multitask learning. MT-Clinical BERT augments the well known BERT \cite{devlin-etal-2019-bert} deep learning architecture with a novel task fine-tuning scheme that allows the learning of features for multiple clinical tasks simultaneously. As a result, our system massively decreases the hardware and computational requirements of deploying BERT into clinical practice by successfully condensing eight 110 million parameter BERT instances into a single model while retaining nearly all BERT-associated task performance gains.
\newpage

Our main contributions are summarized as follows:
\begin{enumerate}
    \item We develop a \textbf{single} deep learning model that concurrently achieves competitive performance over eight clinical tasks spanning named entity recognition, entailment and semantic similarity. As a result, we achieve an eight fold computational speed-up at inference compared to traditional per-task sequentially fine-tuned models.
    \item We demonstrate the feasibility of multitask learning towards developing a universal clinical information extraction system that shares information amongst disjointly annotated datasets.
    \item We release and benchmark against a new and more competitive BERT fine-tuning baseline for eight clinical tasks by performing extensive hyper parameter tuning for each task's dataset. 
\end{enumerate}


\section*{Methods}
This section begins with a description of our clinical multitask learning system and then discusses the clinical text benchmarks evaluating it's performance. In this work, we refer to a task as a tuple consisting of a text dataset and corresponding task objective (e.g. token classification).

\begin{figure}[t!]
  \centering
  \includegraphics[width=\textwidth]{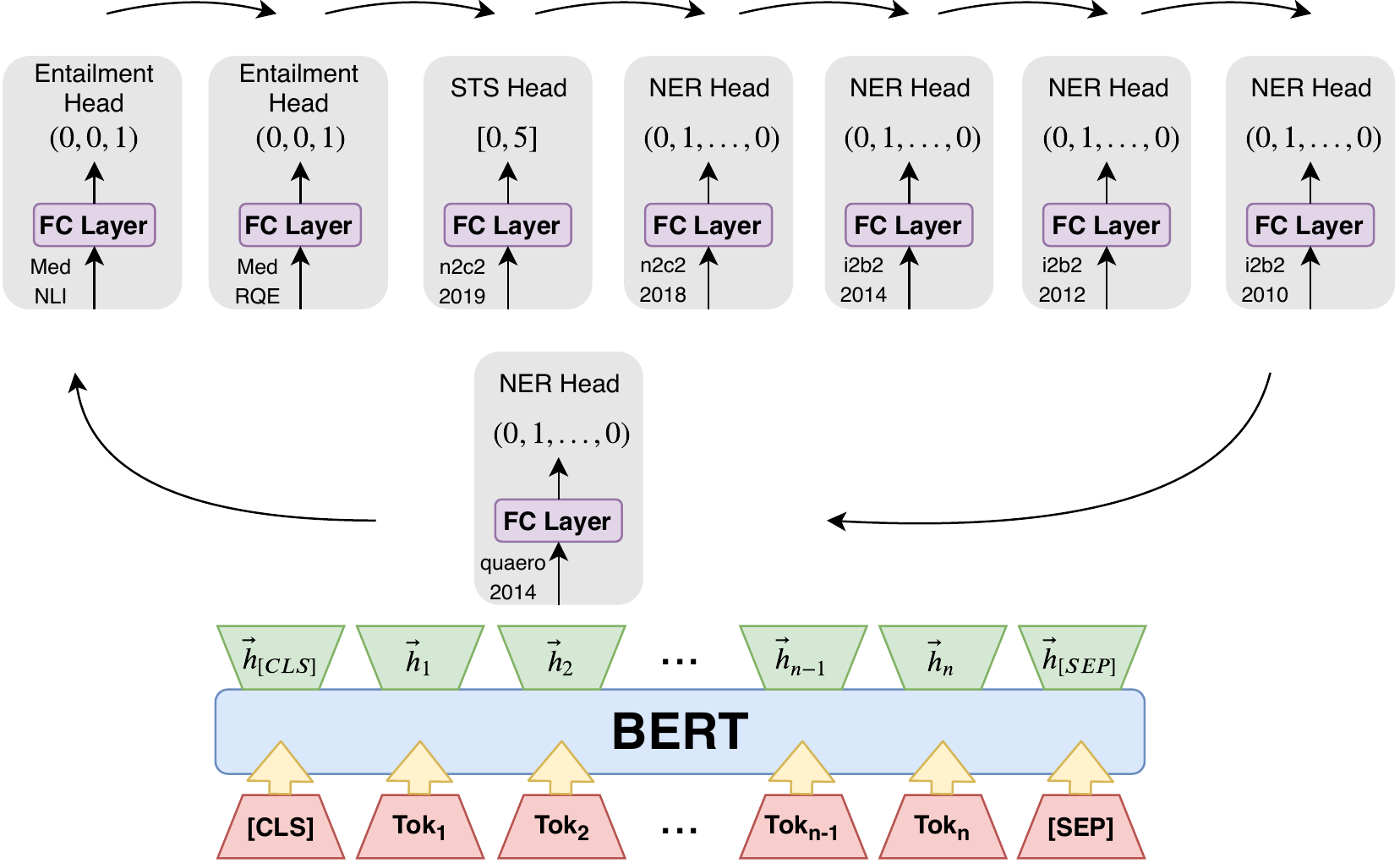}
  \caption{Eight headed MT Clinical BERT with a round robin training schedule. Each Entailment head predicts a one-hot class indicator. The Semantic Text Similarity (STS) head predicts a similarity score in $[0,5]$ representing the semantic similarity of the two input sentences. Each Named Entity Recognition head predicts a one-hot entity indicator for each input sub-word token.}
  \label{fig:architecture}
\end{figure}

\subsection*{Multitasking Clinical BERT}

The standard practice in transfer learning from BERT is a method known as sequential fine-tuning. During sequential fine-tuning, a BERT architecture is initialized with the parameters of a pre-trained, self-supervised model and then fine-tuned with loss signal from a task-specific head. This procedure adapts the weights of the base pre-trained model into a task-specific feature encoder capable of representing the input text such that the task objective is easily discernible by the task-specific head (e.g. linearly separable in the case of classification). In contrast, hard-parameter multitask learning aims to adapt the weights of the base pre-trained model into a feature encoder capable of generating text representations suitable for multiple tasks simultaneously. In the case of BERT, this is achieved by treating the BERT Transformer stack as a feature encoder that feeds into multiple light-weight task-specific architectures each implementing a different task objective.

\begin{figure*}[t!]
    \centering
    \hfill%
    \begin{subfigure}[t]{0.2\textwidth}
        \centering
        \includegraphics[width=.84\textwidth]{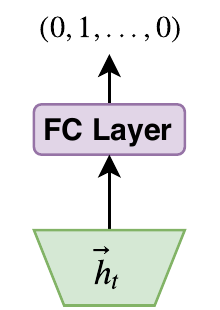}
        \caption{Named Entity Recognition head on a sub-word token hidden state.}
        \label{fig:head_ner}%
    \end{subfigure}\hfill%
    ~
        \begin{subfigure}[t]{0.2\textwidth}
        \centering
        \includegraphics[width=1.1\textwidth]{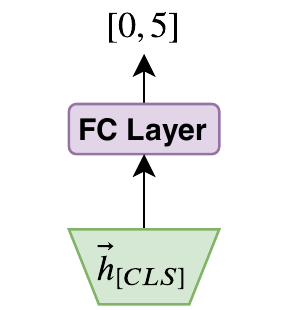}
        \caption{Semantic Text Similarity head on classification token hidden state.}
        \label{fig:head_sts}%
    \end{subfigure}\hfill%
    ~
    \begin{subfigure}[t]{0.2\textwidth}
        \centering
        \includegraphics[width=1.1\textwidth]{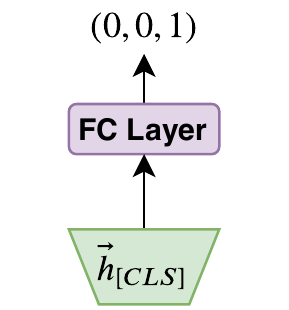}
        \caption{Entailment head on classification token hidden state.}
        \label{fig:head_entailment}%
    \end{subfigure}\hfill%
    \vspace{.2cm}
    \caption{Task-specific heads with corresponding input representations from the BERT hidden state sequence.}
\end{figure*}

Our multitasking model (Figure \ref{fig:architecture}) comprises of a BERT feature encoder with weights initialized from Bio + Clinical BERT \cite{alsentzer-etal-2019-publicly} and eight per-dataset task-specific heads. The head architectures are as follows:
\begin{itemize}
    \item Named Entity Recognition (Figure \ref{fig:head_ner}): token classification via a per-entity linear classifier on sub-word tokens providing loss signal with cross entropy loss.
    \item Semantic Text Similarity (Figure \ref{fig:head_sts}): sentence pair semantic similarity scoring via a linear regression on the sequence representation {[CLS]} token providing loss signal via the mean squared error.
    \item Natural Language Inference (Figure \ref{fig:head_entailment}): sentence pair logical entailment via a linear classifier on the sequence representation {[CLS]} token providing loss signal with cross entropy loss.
\end{itemize}

\begin{algorithm}
	\caption{MT-Clinical BERT Training Schedule} 
	\begin{algorithmic}[1]
	    \Require $\theta_E$: pre-trained Transformer encoder.
    	\Require $\theta_H = \{\theta_{h_1}, \ldots, \theta_{h_n}\}$:  $n$ task-specific heads.
    	
    	\State Randomly initialize $\theta_{h_i}$ $\forall i \in \{1,\ldots,n\}$
    	
    	\While{all batches from largest task dataset are not sampled}
    	\State Sample a batch $D_i$ for each $\theta_{h_i} \in \theta_H$
    	    \For {each $(\theta_{h_i}, D_i)$} \Comment{One round robin iteration}
    	        \State Let $\theta = \theta_{E} \circ \theta_{h_i}$ \Comment{Outputs of encoder into head $\theta_h$}
    	        \State $\theta' = \theta - \alpha \nabla_{\theta} \mathcal{L} \left( \theta, D_i \right)$
    	        \State Update $\theta$ with $\theta'$
    	    \EndFor
    	\EndWhile
	\end{algorithmic} 
\end{algorithm}

To train our multitasking model, the feature encoder must be adapted to support all tasks simultaneously. Their are several established methods of adapting the feature encoder parameters by combining loss signal from each head during training \cite{Ruder2019Neural} (e.g. averaging/adding losses); however, most assume that the loss function is constant across all of the heads. In general this is not necessarily true. When different loss functions are present, the standard  solution is to sub-sample instances from each dataset proportional to the dataset size \cite{Ruder2019Neural} and then proceed with batch stochastic gradient descent with respect to each individual loss function. We propose an alternative training scheme equivalent to proportional sub-sampling, but requiring no additional proportionality calculations. To do this, we cycle the heads and batched gradient updates in a round robin fashion over the BERT feature encoder. This training schedule is summarized in Algorithm 1.


\section*{Data}
In this section, we describe the eight clinical tasks used to evaluate our multitasking system. Table \ref{tab:data} showcases the tasks considered, the pre-defined train and evaluation splits used in our experiments and the corresponding task evaluation metric. 

\begin{table}[h!]
\vspace*{.5cm}
\centering
\caption{Clinical information extraction benchmarks with reported performance metric.}
\label{tab:data}
\begin{tabular}{c|c|c|c|c|c}
Task & Dataset & Metric & Description & \# Train Inst.& \# Test Inst.\\
\hline
STS & n2c2-2019 \cite{N2C2_2019} & Pearson Rho & Sentence Pair Semantic Similarity & 1,641 & 410 \\
\hline
\multirow{2}{*}{Entailment} 
&    MedNLI \cite{romanov2018lessons} & Accuracy & Sentence Pair Entailment & 12,627 & 1,422\\
&   MedRQE \cite{RQE:AMIA16} & Accuracy & Sentence Pair Entailment & 8,588 & 302\\    
\hline
\multirow{5}{*}{NER} 
&   n2c2-2018 \cite{henry20202018} & Micro-F1  & Drug and Adverse Drug Event & 36,384 & 23,462 \\    
&   i2b2-2014 \cite{stubbs2015automated} & Micro-F1  & PHI de-identification & 17,310 & 11,462 \\
&   i2b2-2012 \cite{sun2013evaluating}   & Micro-F1  & Events & 16,468 & 13,594\\    
&   i2b2-2010 \cite{uzuner20112010} & Micro-F1  & Problems, Treatments and Tests & 27,837 & 45,009 \\
&   quaero-2014 \cite{neveol14quaero} & Micro-F1 & UMLS Semantic Groups (French) & 2,695 & 2,260\\
\hline
\end{tabular}
\end{table}

The Semantic Textual Similarity (STS) task is to assign a numerical score to sentence pairs indicating their degree of semantic similarity. Our system includes one STS dataset: 
\begin{enumerate}
    \item[1.] The n2c2-2019 dataset consists of de-identified pairs of clinical text snippets from the Mayo Clinic that were ordinally rated from 0 to 5 with respect to their semantic equivalence where 0 indicates no semantic overlap and 5 indicates complete semantic overlap.  The training dataset contains 1,642 sentence pairs; while the test dataset contains 412 sentence pairs. 
\end{enumerate} 
Textual Entailment is the task of determining if one text fragment is logically entailed by the previous text fragment. We utilize two Entailment datasets: 
\begin{enumerate}
    \item[2.] The MedNLI\cite{romanov2018lessons} dataset consists of the sentence pairs developed by Physicians from the Past Medical History section of MIMIC-III clinical notes annotated for {\it Definitely True}, {\it Maybe True} and {\it Definitely False}. The dataset contains 11,232 training, 1,395 development and 1,422 test instances. We combined the training and development instances for our work. 
    
    \item[3.] The MedRQE\cite{RQE:AMIA16} dataset consists of question-answer pairs from the National Institutes of Health (NIH) National Library of Medicine (NLM) clinical question collection consisting of Frequently Asked Questions (FAQs). The positive examples were drawn explicitly from the dataset while the negative pairs were collected by associating a randomly combined question-answer pair as having at least one common keyword and at least one different keyword from the original question. The dataset contains of 8,588 training pairs and 302 test pairs with approximately 54.2\% as positive instances.
\end{enumerate}

Named Entity Recognition (NER) is the task of automatically identifying mentions of specific entity types within unstructured text. In this work, we utilize five NER datasets: 
\begin{enumerate}
    \item[4.] The n2c2 2018 dataset\cite{henry20202018}  consists of 505 de-identified discharge summaries drawn from the MIMIC-III clinical care database and annotated for Adverse Drug Events (ADEs) and the drug that caused them; reason for taking the drug and the associated dosage, route, and frequency information. The training and test sets contain 303 and 202 instances respectively.  

    \item[5.] The i2b2-2014 dataset\cite{sun2013evaluating} consists of 28,772 de-identified discharge summaries provided from Partners HealthCare annotated for personal health information (PHI) including, patient names, physician names, hospital names, identification numbers, dates, locations and phone numbers. The training and test sets contain 17,310 and 11,462 instances respectively. 

    \item[6.] The i2b2-2012 dataset consists of de-identified discharge summaries provided by Partners HealthCare and MIMIC-II.  The dataset was annotated for two entity types: 1) clinically events, including both clinical concepts, departments, evidentials and occurrences; and (2) temporal expressions, referring to the dates, times, durations, or frequencies. In this work, we evaluated over only the event annotations. The training and test sets contain 16,468 and 13,594 instances respectively. 

    \item[7.] The i2b2-2010 dataset\cite{uzuner20112010} consists of de-identified discharge summaries provided by Partners HealthCare and MIMIC-II; and de-identified discharge and progress notes from the University of Pittsburg Medical Center. The dataset was annotated for three entity types: 1) clinical concepts, clinical tests and clinical problems. These entities overlap with the i2b2-2010 event annotations. The training and test sets contain 27,837 and 45,009 instances respectively. 

    \item[8.] The quaero-2014 dataset \cite{neveol14quaero} consists of a french medical corpus containing three document types: 1) the European Medicines Agency (EMEA) drug information; 2) MEDLINE research article titles; and 3) European Patent Office (EPO) patents. The dataset was annotated for ten types of clinical entities from the Unified Medical Language System (UMLS) Semantic Groups\cite{mccray2001aggregating}: Anatomy, Chemical and Drugs, Devices, Disorders, Geographic Areas, Living Beings, Objects, Phenomena, Physiology, Procedures.The training and test sets contain 2,695 and 2,260 instances respectively. 

\end{enumerate}

\section*{Evaluation}

To insure a competitive and fair comparison with existing state-of-the-art solutions, we perform a hyper parameter search for each individual Clinical BERT task fine-tuning run and report the best performing model on each task. Recent work \cite{dodge2020finetuning} has found negligible performance differences between random seed re-initialization and more complex methods of hyper parameter search during BERT fine-tuning so we opt for the former. Specifically, for each task a Clinical BERT instance is initialized and fine-tuned for twenty training data epochs over five unique random seeds resulting in 100 unique task-specific models. We report the top performing model at evaluation. We do not utilize a development set for training MT-Clinical BERT as the multitasking paradigm itself largely removes the ability for a model to overfit any specific task. Additionally, we do not perform hypothesis testing due to the significant computational resources required.

\section*{Results and Discussion}

\begin{table}[htp]
\vspace*{.5cm}
\centering
\caption{Clinical information extraction performance of MT Clinical BERT versus hyper parameter searched Clinical BERT fine-tuning runs. All span level metrics are exact match. Task performances showcased in the column \textit{MT-Clinical BERT} represent a \textbf{single} multitask trained feature encoder with individual task-specific heads. All other reported results are generated from task-specific BERT models. Higher is better.}
\label{tab:performance_results}
\begin{tabular}{c|c|c|c}

& MT-Clinical BERT & Optimized Clinical BERT & Clinical Bert \cite{alsentzer-etal-2019-publicly} \\
\hline
n2c2-2019 & 86.7 $(-0.5)$ & 87.2 & -\\
MedNLI & 80.5 $(-2.3)$ & 82.8 & 82.7\\
MedRQE & 76.5 $(-3.6)$  & 80.1 & - \\
n2c2-2018  & 87.4 $(-0.7)$  & 88.1 & -\\
i2b2-2014  & 91.9 $(-3.6)$ & 95.5 & 92.7\\
i2b2-2012  & \textbf{84.1 $(+0.2)$}  & 83.9 & 78.9\\
i2b2-2010 & 89.5 $(-0.3)$ & 89.8 & 87.8\\
quaero-2014 & 49.1 $(-6.4)$ & 55.5 & -\\
\hline
\end{tabular}
\end{table}

We compare the performance of our multitasking information extraction system to state-of-the-art BERT sequential fine-tuning baselines in Table \ref{tab:performance_results}. Evaluations reported in the columns \textit{(Optimized) Clinical BERT} represent individually fine-tuned, per-task BERT models. Evaluations reported in the column \textit{MT-Clinical BERT} represent light-weight task-specific heads over a single multitask trained BERT feature encoder. We find that the performances reported in the Clinical BERT \cite{alsentzer-etal-2019-publicly} paper can be substantially improved via hyper parameter search. While this is not surprising (the authors specify that performance was not their goal), it is important to compare improvements or degradations against a competitive baseline. All further discussion compares the multitasking model to the hyperparameter \textit{Optimized Clinical BERT} baseline.

We observe a slight but consistent performance degradation in MT-Clinical BERT relative to sequential finetuning. Intuitively, this suggests that learning a general clinical text representation capable of supporting multiple tasks has the downside of losing the ability to exploit dataset or clinical note specific properties when compared to a single, task-specific model. This phenomena can best be illustrated amongst the English token classification tasks, where the de-identification task, i2b2-2014, suffered the greatest performance degradation. 
Clinical BERT is pretrained over MIMIC-III. As MIMIC-III is de-identified, all PHI markers in the original notes are replaced with special PHI tokens that do not linguistically align with the surrounding text (e.g. an instance of a hospital name would be replaced with the token [HOSPITAL]). Due to this, no PHI tokens are present in MIMIC-III and thus the pre-training procedure of Clinical BERT over the MIMIC-III corpus provides little signal pertaining to PHI tokens. Alsentzer et al.\cite{alsentzer-etal-2019-publicly} observes and discusses this property at depth. These results suggest that a lack of PHI related information during pre-training can be overcome by the encoder during sequential fine-tuning but not as successfully when regularized by the requirement of supporting multiple tasks.

Surprisingly, MT-Clinical BERT confers a slight performance increase in the problem, treatment and test extraction task i2b2-2012 relative to the hyper parameter tuned Clinical BERT baseline. This suggests that multitask regularization with the related problem, treatment and test extraction task in i2b2-2010 may be inducing features more suited to generalizability for these entity types. These are the only NER tasks with overlapping entity definitions.

Our final observation re-enforces the commonly laid out claim in the multitasking community related to task orthogonality / overlap. In the supervised multitask set-up, two tasks are said to have overlap when some characteristics of a given task (e.g. data domain, task objective, target label space, etc) should intuitively help with performance on a different but related task. Otherwise, tasks are said to be orthogonal along that characteristic \cite{Ruder2019Neural}. The majority of the tasks $(5/8)$ in this study are token classification objectives. Unlike the three segment level tasks, these require the BERT feature encoder to learn task-robust contextual token representations which, due to their prevalence during training, may negatively harm the formation of segment level representations. This objective orthogonality is suggested by consistent and large performance decreases in the entailment tasks (MedNLI and MedRQE). We speculate that this could be aided by including additional clinical-related segment level objectives during training or by incorporating the original next sentence prediction pre-training objective into the multitasking mix. Similarly, the quaero 2014 corpus is entirely in French. This naturally induces a lingual orthogonality relative to the other seven English corpora. This orthogonality manifests by inducing the largest loss in competitiveness (-6.4\%) to fine-tuning baselines across all tasks. Again, we suspect that the inclusion of additional non-English token level tasks could close this performance gap.

To summarize, the main insights from our analysis are:
\begin{itemize}
    \item A general trend of degradation in MT-Clinical BERT task-specific performance over individual task-specific models. This is a direct trade-off to the eight-fold reduction in parameters and computational speed up at inference provided by MT-Clinical BERT.
    \item The observation of the task-specific performance increase on i2b2-2012 by MT-Clinical BERT. This is  potentially due to the the regularization provided during multitask learning.
    \item The observation that the greatest relative reduction in multitasking performance occurs on datasets (MedNLI, MedRQE and quaero-2014) with orthogonal characteristics to the predominately English token classification (NER) tasks considered.
\end{itemize}

\section*{Experimental Details and Reproducibility}
We base our implementation on the well-known HuggingFace ``Transformers'' implementation of BERT. During hyper parameter tuning, we re-initialize with random seeds in the set $\{1,\ldots, 5\}$. All fine-tuning is performed with constant learning rate $5e^{-5}$. The NER heads train with 512 sub-word sequences of batch size 25, while the STS and Entailment training is performed with a batch size of 40. All training and evaluation is conducted on a single Tesla V100 GPU. In addition to our pre-trained models, we support reproducibility by including all pre-processing necessary to replicate our results in the code release.

\section*{Avenues for Practical Impact}
It is important that the recent NLP advances our work builds on can reach clinical practice. This section provides insight to the clinical NLP practitioner regarding the feasibility and advantages of transitioning MT-Clinical BERT into their clinical EHR analysis systems. The main contribution of this work is condensing eight 110M parameter deep learning models into \textbf{one single} 110M parameter model while largely retaining the performance gains contributed by recent NLP advances. Practically, this means that clinical note engineers can experience \textit{enormous computational benefits} (8x) at inference while performing \textit{less implementation work} by implementing a clinical information system based on our contribution. Importantly, our contribution is expandable via the integration of additional tasks during training. This means our system is capable of integrating and concurrently supporting future information extraction tasks such as the inclusion of novel, currently undefined named entity types.

\section*{Limitations}
We foresee the following limitations for both the implementation and scaling of our proposed system. First, the datasets considered are annotated over patient discharge summaries. Naturally, different types of notes may have differing underlying data distribution which can lead to performance degradation. Second, we have observed from experiments in other domains that scaling the number of tasks $(>40)$ during training inversely correlates with per-task performance. This means that multitask training with a large number of tasks may require careful ablation experiments to gauge the net benefit of adding any given task.

\section*{Related Work}
Multitask learning has been an integral sub-field of the machine learning community for many decades. In the context of deep learning, programs in several domains spanning drug discovery, computer vision and natural language processing have continued achieving successes by sharing supervised signal and data between machine learning tasks. Half a decade ago, Ramsdunar et al. \cite{ramsundar2015massively} introduced a multitask learning system based on hard-parameter sharing for drug discovery. This contribution achieved significant performance improvement in drug target identification by leveraging 259 unique drug target tasks during multitask training. Similarly in computer vision Yan et al. \cite{Yan2019MULANMU} developed MULAN, a multitasking system that concurrently detects, tags and segments lesions in radiological report images. In 2019, Liu et al.\cite{liu-etal-2019-multi-task} introduced a multitasking model to improve performance on the GLUE natural language understanding benchmark and Google introduced T5 \cite{raffel2019exploring} - a system capable of multitasking by framing common NLP tasks as sequence generation objectives sharing a single encoder/decoder.

\section*{Future work}

There are several directions for future work. We describe them and provide insight below.

\begin{itemize}
    \item \textbf{Adding more tasks and datasets.} Is adding more tasks feasible and beneficial?
    Their is strong evidence \cite{ramsundar2015massively} suggesting that including a greater number of overlapping tasks may increase task-specific predictive performance. This comes with the additional benefit of increasing computational performance at inference as described in this work.
    \item \textbf{Learning from limited data.} Do the representations obtained via multitask learning serve as a better initialization for learning from limited data resources? Work in this direction would benefit from the inclusion of instance ablation studies.
    \item \textbf{Unifying NLP pipelines into end-to-end systems.} Many common NLP tasks build upon the output of previous tasks. This naturally results in phenomena such as error propagation. Can the shared representations produced by a multitask encoder construct an effective joint NER and relation identification system? Recent work \cite{giorgi2019endtoend} suggests this is possible but can it be accomplished in the multitasking framework?
    \item \textbf{Incorporating pre-training objectives during multitasking.} In low annotated data domains such as clinical text, we suspect it may be useful to incorporate the self-supervised masked language modeling and next sentence prediction objectives during multitask training. During preliminary experiments, we find that this does not harm system performance.
\end{itemize}

\section*{Conclusion}

We find that multitask learning is an effective mechanism to distill information from multiple clinical tasks into a single system. This has the main benefit of significant hardware and computational reductions at inference with the trade-off of a small performance degradation. Our system directly increases the potential for the use of recent state-of-the-art NLP methods in clinical application. In addition, we contribute new state-of-the-art baselines for several clinical information extraction tasks. The data repositories and resources of the clinical NLP community have grown steadily over the past two decades - the doors have been opened to consolidate, cross-leverage and jointly build on these expensive annotation efforts. We make our implementation and pre-trained models publicly accessible \footnote{\url{https://github.com/AndriyMulyar/multitasking\_transformers}}.

\section*{Acknowledgements}
The authors would like to thank Nick Rodriguez for his valuable commentary and suggestions on the final draft of this article.





\makeatletter
\renewcommand{\@biblabel}[1]{\hfill #1.}
\makeatother

\bibliography{main}
\bibliographystyle{unsrt}



\end{document}